%% file: main.tex
\documentclass[10pt,twocolumn,letterpaper]{article}

\usepackage{graphicx}
\usepackage{amsmath}
\usepackage{amssymb}
\usepackage{booktabs}
\usepackage{multirow, booktabs}
\usepackage{float}
\usepackage{xcolor}
\usepackage{subcaption}
\usepackage[pagebackref,breaklinks,colorlinks]{hyperref}
\usepackage[capitalize]{cleveref}
\crefname{section}{Sec.}{Secs.}
\Crefname{section}{Section}{Sections}
\Crefname{table}{Table}{Tables}
\crefname{table}{Tab.}{Tabs.}

\def\methodabbrev short{ActionDiffusion}
\def\method full{\textbf{Action}-aware Noise Mask \textbf{Diffusion}}

\begin{document}

\title{\methodabbrev short: An Action-aware Diffusion Model for Procedure Planning in Instructional Videos}

\author{Lei Shi\\
Institute for Visualisation and Interactive Systems, University of Stuttgart, Germany\\
{\tt\small lei.shi@vis.uni-stuttgart.de}
\and
Paul Bürkner\\
Department of Statistics, TU Dortmund University\\
{\tt\small paul.buerkner@gmail.com}
\and 
Andreas Bulling\\
Institute for Visualisation and Interactive Systems, University of Stuttgart, Germany\\
{\tt\small andreas.bulling@vis.uni-stuttgart.de}
}
\date{} 
\maketitle

\begin{abstract}
    \input{Sections/abstract}

\end{abstract}

\input{Sections/introduction}

\input{Sections/related_work}

\input{Sections/method}
\input{Sections/experiments}

\input{Sections/results}

\input{Sections/conclusion}

{\small
\bibliographystyle{ieee_fullname}
\bibliography{main}
}

\end{document}

%% file: Sections/abstract.tex
We present \textit{\methodabbrev short} -- a novel diffusion model for procedure planning in instructional videos that is the first to take temporal inter-dependencies between actions into account.
Our approach is in stark contrast to existing methods that fail to exploit the rich information content available in the particular order in which actions are performed.
Our method unifies the learning of temporal dependencies between actions and denoising of the action plan in the diffusion process by projecting the action information into the noise space.
This is achieved 1) by adding action embeddings in the noise masks in the noise-adding phase and 2) by introducing an attention mechanism in the noise prediction network to learn the correlations between different action steps.
We report extensive experiments on three instructional video benchmark datasets (CrossTask, Coin, and NIV) and 
show that our method outperforms previous state-of-the-art methods on all metrics on CrossTask and NIV and all metrics except accuracy on Coin dataset. We show that by adding action embeddings into the noise mask the diffusion model can better learn action temporal dependencies and increase the performances on procedure planning.

%% file: Sections/introduction.tex
\section{Introduction}
\label{sec:intro}

To support humans in everyday procedural tasks, such as cooking or cleaning, future autonomous AI agents need to be able to plan actions from visual observations of human actions and their environment -- so-called procedure planning \cite{lippi2020latent,chang2020procedure}.
Procedure planning is commonly defined as the task of predicting an action plan, i.e., a sequence of individual actions, from only a start and final observation of the overall procedure.
Several previous works have investigated procedure planning from visual observations \cite{han2020deep, van2021deepkoco, hafner2019learning}.
But these works have learnt visual representations from artificial and rather simple images, such as a simulated cart pole.
Other works have used real-world images to learn action plans \cite{fang2019dynamics, sun2022plate} but the environment was simplified and constrained by pre-defined object-centred representations \cite{kumar2023learning}, for example, coloured cubes as objects on a table.
Leveraging more advanced deep learning methods for planning procedures from instructional videos has the potential to address this limitation \cite{chang2020procedure, bi2021procedure, zhao2022p3iv}. Although different approaches have been developed to tackle the procedure planning task, the challenge remains open due to the complex and unstructured video observations.

\begin{figure}[t]
    \centering
    \includegraphics[width=\linewidth]{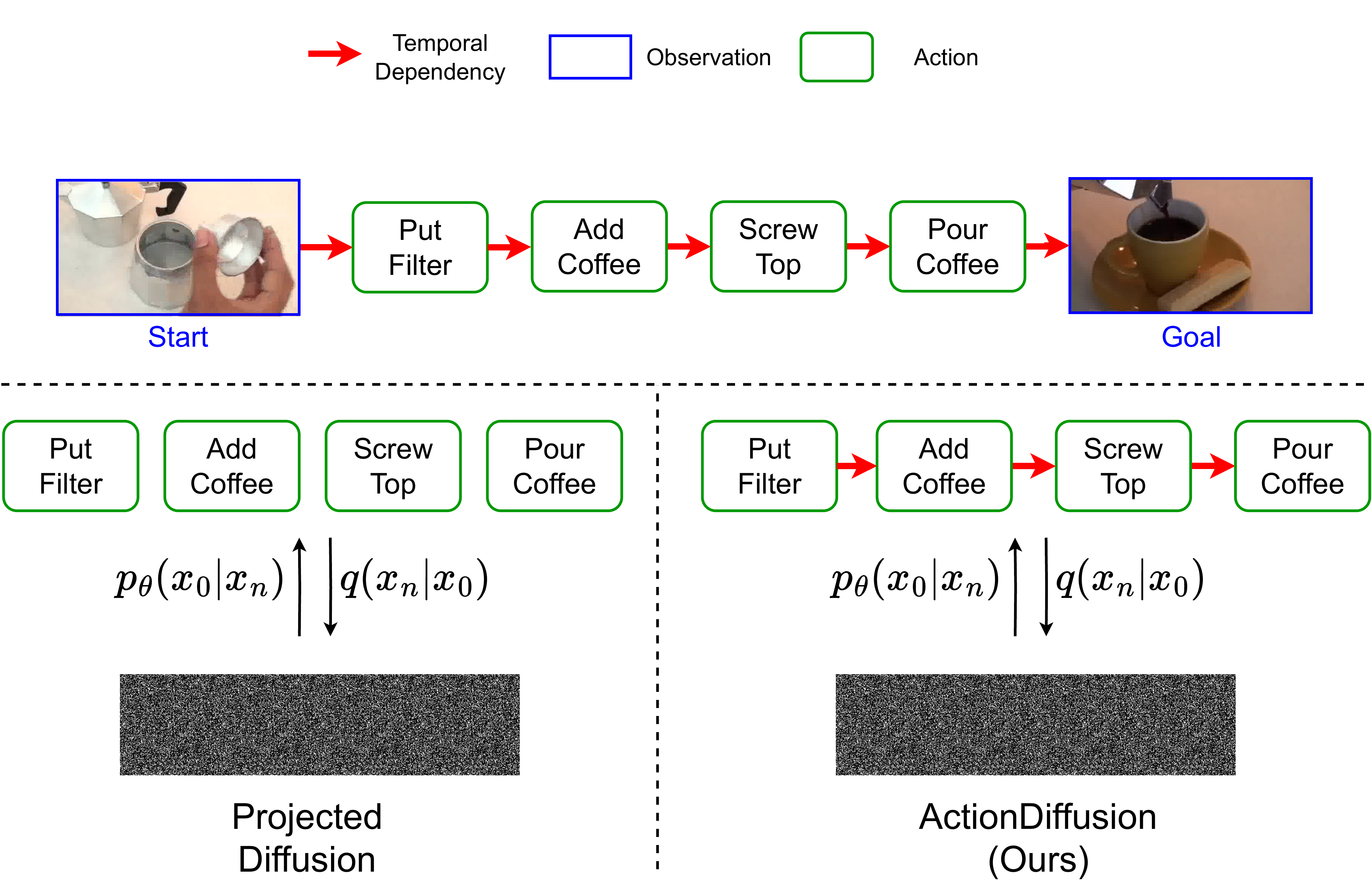}
    \caption{Procedure planning in instructional videos using diffusion models. \textbf{Upper section:} Procedure planning task is to generate intermediate actions given the start and goal observation. \textbf{Lower left section:} Previous work (Projected Diffusion) \cite{wang2023pdpp} does not take the temporal dependencies between actions into account. \textbf{Lower right section:} Our method incorporates these dependencies into the diffusion model.}
    \label{fig:teasor}
\end{figure} 

Diffusion models have achieved outstanding results in many research fields such as image generation \cite{ho2020denoising, nichol2021improved}, text-to-image generation \cite{rombach2022high, zhang2023adding, ruiz2023dreambooth}, trajectory planning \cite{janner2022planning, huang2023diffusion}, video generation \cite{ho2022imagen, luo2023videofusion}, human motion prediction \cite{yan2023gazemodiff}, time series imputation and generation \cite{tashiro2021csdi, jiao2024diffgaze} and so on. 
The current state-of-the-art method for procedure planning in videos \cite{wang2023pdpp} is also based on a diffusion model. 
Unlike the diffusion models for images, the input for the diffusion model is a multi-dimensional matrix that consists of the visual observations of the start and goal, the sequence of actions, and task classes. At inference time, the action plan is taken from the action sequence in the generated full matrix. 
A key limitation of this work is that the influence of the temporal dependencies of actions, i.e. that actions are more likely to cause particular follow-up actions, has not been considered. Although the input matrix in \cite{wang2023pdpp} contains the action labels during the training, the diffusion model still treats the input matrix as a static ``image'', it does not learn these temporal dependencies. 

To overcome this limitation we propose \method full (\textbf{\methodabbrev short)} --
the first method that leverages the temporal dependencies between actions into the diffusion process
(see \autoref{fig:teasor}).
Our method learns the temporal dependencies of actions in the noise space of the diffusion instead of the feature space, 
thereby unifying the tasks of learning temporal dependencies and generating action plans in the diffusion steps.
More specifically, we first propose an action-aware noise mask for the noise-adding stage of the diffusion model. 
We add the action embeddings in addition to the Gaussian noise in the noise masks so that the model input is transformed into Gaussian noises by iteratively adding the action-aware noise. 
Second, we introduce an attention mechanism in the denoising neural network (U-Net) to learn the correlations between actions. During the inference, it can then predict action-aware noises to generate action plans.
In line with previous work \cite{zhao2022p3iv,wang2023pdpp}, we evaluate the performance of \methodabbrev short through experiments on three popular instructional video benchmarks: CrossTask \cite{zhukov2019cross}, Coin \cite{tang2019coin}, and NIV \cite{alayrac2016unsupervised} with various time horizons.
Our results show that \methodabbrev short achieves State-Of-The-Art (SOTA) performances across different metrics for all time horizons on all three datasets. We demonstrate that by adding action embeddings into the noise mask, the diffusion model can effectively learn the temporal dependencies between actions and increase the performance on procedure planning in instructional videos.

The specific contributions of our work are threefold:
1) We propose \methodabbrev short, a novel method incorporating the action temporal dependencies in the diffusion model for procedure planning in instructional videos. We unify the learning of temporal dependencies and action plan generation in the noise space.
2) We add action embeddings into noise masks in the noising-adding stage of diffusion models and use a denoising neural network with self-attention to better learn and predict the action-aware noise to reconstruct the action plan in the denoising phase. 
3) We evaluate our methods on CrossTask, Coin, and NIV datasets across various time horizons and achieve SOTA performances in multiple metrics and show the advantage of incorporating action temporal dependencies in the diffusion model, which previous work did not consider.

%% file: Sections/related_work.tex
\section{Related Work}
\label{sec:related_work}

\subsection{Learning Actions from Videos}
There are several lines of research in learning actions from videos. Action recognition \cite{wang2018temporal, yang2020temporal} is to classify what actions humans are performing in videos. This is a video classification problem. Action anticipation \cite{furnari2020rolling, gong2022future} is the task of predicting future actions based on the video given to a model. 
The task of procedure planning differs from action recognition and action anticipation, it plans the action sequence between the given visual input of the start and goal state. 
There were works to learn and plan actions from visual input \cite{han2020deep, van2021deepkoco, hafner2019learning}. However, these works learnt visual representations from rather simple simulated images. Other works used real-world images to learn action plans \cite{fang2019dynamics, sun2022plate}, however, the environment was still simple and constrained. Understanding the natural real-life scenario in which humans are present to plan actions is still a challenge and it requires models to have the ability to understand complex visual scenes and human actions.

\subsection{Procedure Planning}
The task of procedure planning was defined by \cite{chang2020procedure}. The authors used CNNs to extract visual features and modeled the dynamics between observations and actions in the feature space using Multi-Layer Perceptrons (MLPs) and Recurrent Neural Networks (RNNs). In \cite{sun2022plate}, the authors followed the same paradigm of modelling the dynamics between actions and observations but used transformers instead. RL was used in \cite{bi2021procedure} to learn the action policy. All these works used a separate planning algorithm (Beam search or Walk Through) during planning. Later works used generative models to sample action plans during the inference stage. In \cite{zhao2022p3iv}, a generative component together with transformers was trained for sampling action plans during inference. 
The denoising diffusion probabilistic model was used for procedure planning in \cite{wang2023pdpp}. The action plan, task class, $o_S$ and $o_g$ were formed as the input to the diffusion model and the action plan was denoised from random noise.

\subsection{Modeling Temporal Dependencies}
Previous works used different approaches to model the temporal dependencies between actions in procedure planning. In \cite{chang2020procedure}, the authors modeled the temporal dependencies based on Markov Decision Process (MDP), where the action at $t+1$ is based on the current state $x_t$ and the current action $a_t$. As transformers \cite{vaswani2017attention} have shown strong capabilities in sequence modeling, 
P3IV \cite{zhao2022p3iv} used the transformer to empirically model the temporal dependencies by putting the start and goal observations as the first and last queries in the transformer and making the intermediate queries (related to action sequences) learnable. PlaTe \cite{sun2022plate} also used MDP for modeling temporal dependencies, specifically two transformers, i.e. an action transformer and a state transformer were used. During training, all actions and states were used and during inference all past action-state pairs were used.  Although we also use self-attention to model the temporal dependencies, which is a key component in transformers, there are two differences. First, we build the correlation between actions by accumulating action embeddings in noise-adding stage in the diffusion model. Second, we add self-attention in the U-Net to enhance the temporal relations during denoising.

%% file: Sections/method.tex
\section{Method}
\label{sec:method}

\begin{figure*}[t]
    \centering
    \includegraphics[width=0.8\textwidth]{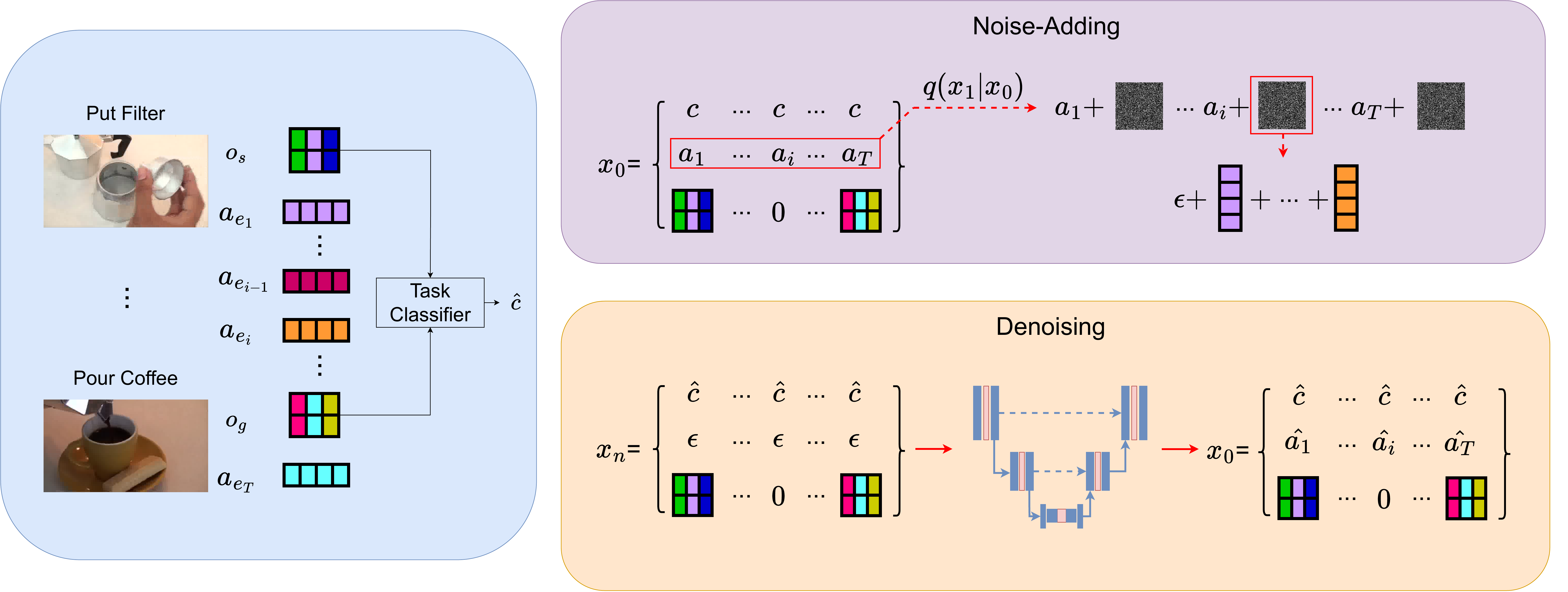}
    \caption{Overview of \methodabbrev short. From an instructional video, we extract the visual feature of the start state $o_s$ and the goal state $o_g$ as well as the features of actions $a_{e_{1:T}}$. We use the task class $c$, one-hot action class $a_{1:T}$, $o_s$ and $o_g$ as the input of the diffusion model. Note that in the training, we use the ground truth task class $c$ and predicted task class $\hat{c}$ during inference. A separate task classifier is trained to get $\hat{c}$. In the noise-adding phase in training, the noise is added on $a_{1:T}$. For each action, we add all previous action embeddings and the current action embedding in addition to the Gaussian noise. In the denoising phase during inference, we use the U-Net with attention to predict the action-aware noise to denoise $x_n$. The predicted action plan is the action sequence $\hat{a}_{1:T}$ from the reconstructed input $x_0$.}
    \label{fig:method_overview}
\end{figure*}

\subsection{Problem Formulation}
\label{subsec:problem_form}
We adopt the problem formulation of procedure planning used in previous work \cite{chang2020procedure, wang2023pdpp}:
Given the visual observation of the start state $o_s$ and the goal state $o_g$, the task is to predict the intermediate steps, i.e., the action plan $\pi=a_{1:T}$ for a chosen time horizon $T$.
The action plan $\pi$ will transform $o_s$ to $o_g$. 
More formally, procedure planning can be written as,
\begin{equation}\label{eq:problem_formation}
    p(\pi \mid o_s, o_g) = \int p(\pi \mid o_s, o_g, \hat{c}) \; p(\hat{c} \mid o_s, o_g) \, d\hat{c},
\end{equation}
where $\hat{c}$ is the predicted task class of the video (e.g., make coffee) and to complete a task, a sequence of actions needs to be performed.
The planning is decomposed into two steps \cite{wang2023pdpp}: 1) predicting the task class $\hat{c}$ given the start state $o_s$ and the goal state $o_g$, 2) inferring the action plan $\pi$ given $o_s$, $o_g$, and $\hat{c}$ by sampling from the diffusion model.

\subsection{Action-aware Noise Mask Diffusion}
\autoref{fig:method_overview} provides an overview of our proposed \methodabbrev short method.
Taking an instructional video as input, the method first extracts the visual features of the start state $o_s$, the goal state $o_g$, and of action embeddings $a_{e_{1:T}}$. 
It then uses the task class $c$, one-hot action class $a_{1:T}$, $o_s$, and $o_g$ to form the input for the diffusion model. 
Note that during training, we use the ground truth task class $c$ while the predicted task class $\hat{c}$ is used during inference. 
To obtain $\hat{c}$ we train a separate task classifier that models $p(\hat{c} \mid o_s, o_g)$ in \autoref{eq:problem_formation}. 
In the noise-adding phase, during training, the noise is added on $a_{1:T}$ in the model input $x_0$. 
For each action, we add all previous as well as the current action embedding in addition to the Gaussian noise. 
In the denoising phase, during inference, we use a U-Net with attention to predict the action-aware noise to denoise the noisy input $x_n$ at step $n$. The action sequence $\hat{a}_{1:T}$ from the reconstructed input $x_0$ is the predicted action plan (see below for details).

\subsection{Diffusion Model}

A diffusion model \cite{ho2020denoising} takes input $x_0$ and performs two steps on the input: 
The first one is the noise-adding step, where Gaussian noise $\epsilon \sim \mathcal{N}(0, \mathbf{I})$ is added to $x_0$ incrementally and eventually $x_0$ approaches a standard Gaussian distribution in $x_N$. This noise-adding process $q(x_n \mid x_{n-1})$ for $n = N, \ldots, 1$ is described by the following equation,
\begin{equation}
    q(x_n \mid x_{n-1})=\mathcal{N}(x_n;\sqrt{1-\beta_n}x_{n-1}, \beta_n\mathbf{I}), \label{eq:q_sample}
\end{equation} 
where $\beta_n \in (0,1)$ is pre-defined (see below). $\beta_n$ decides how much of the noise is added to $x_n$. A re-parameterization is then applied,
\begin{equation}\label{eq:repara}
    x_n = \sqrt{\bar{\alpha}_n}x_0 + \sqrt{1-\bar{\alpha}_n}\epsilon,
\end{equation}
where $\bar{\alpha}_n = \prod_{s=1}^n(1-\beta_s)$. A cosine noise scheduling technique \cite{nichol2021improved} is used to determine $\{\beta_s\}_{s=1}^n$, 
\begin{equation}
    \bar{\alpha}_n = \frac{f(n)}{f(0)}, \\
    f(n) = \text{cos}(\frac{n/N+\tau}{1+\tau}\times\frac{\pi}{2})^2,
\end{equation}
where $\tau$ is an offset value to prevent $\beta_n$ from becoming too small when $n$ is close to 0. 

The second step is the denoising process, where the diffusion model samples $x_N$ from Gaussian noise $\mathcal{N}(0, \mathbf{I})$ and denoises $x_N$ to obtain $x_0$ via the denoising process
\begin{equation}\label{eq:unet}
    p_\theta(x_{n-1} \mid x_n) = \mathcal{N}(x_{n-1};\mu_\theta(x_n,n),\Sigma_\theta(x_n,n)),     
\end{equation}
where $\mu_\theta(x_n,n)$ is parameterised by a neural network $\epsilon_\theta(x_n,n)$, and $\Sigma_\theta(x_n,n)$ is calculated by using $\beta_n\mathbf{I}$.

\subsection{Action-aware Noise Mask}
We follow \cite{wang2023pdpp} to construct the input $x_0$ to the diffusion model as,
\begin{equation}
x_0 = 
    \begin{bmatrix}
        c & c & ... & c & c \\
        a_1 & a_2 & ... & a_{T-1} & a_T \\
        o_s & 0 &... &0 & o_g
    \end{bmatrix},
\end{equation}
at the noise-adding step, the noise is only added to the action dimension, i.e. $a_{1:T}$ and $a_i$ is a one-hot vector of a number of $A$ action classes, thus $\mathbf{I} \in \mathbb{R}^{T\times A}$ in Equation~\eqref{eq:q_sample} and $x_0 \in \mathbb{R}^{T\times (O+A+C)}$, $O$ and $C$ are the dimension of $o_s$ and the dimension of task class $c$. We design an action-aware noise mask $M_a$ with multiple previous actions accumulated (MultiAdd) as
\begin{equation}\label{eq:multi-add_mask}
M_a = 
    \begin{bmatrix}
        0 & 0 & ... & 0 \\
        g(a_{e_{1}}) & \sum_{i=1}^2g(a_{e_{i}}) & ...  & \sum_{i=1}^{T}g(a_{e_{i}}) \\
        0 & 0 & ... &0 
    \end{bmatrix},
\end{equation}
where $a_{e_{1:T}}$ are the embeddings of $a_{1:T}$, and $g(\cdot)$ normalises the values of action embeddings to the range of [-1, 1]. $M_a$ has the same dimension as $\mathbf{I}$. Since the denoising step clamps the feature to the range of [-1, 1] too, we normalise the action embeddings for more stable training. 
In addition to the Gaussian noise applied to $x_0$, we add the normalised action embeddings to the noise mask. In the temporal direction, we accumulate all previous action embeddings. Our intuition is that the actions will affect the states and this should be reflected in the noise space. At each $i\in[2:T]$ the noise mask knows what the previous actions are and $\mu_\theta$ can learn the temporal dependencies of actions in the denoising stage. 
With the action-aware noise mask, \autoref{eq:q_sample} then becomes,
\begin{equation}\label{eq:q_sample_act}
    q(x_n \mid x_{n-1})=\mathcal{N}(x_n;\sqrt{1-\beta_n}x_{n-1}, \beta_n (\mathbf{I}+M_a)), 
\end{equation} 
we again apply re-parameterisation and update \autoref{eq:repara},
\begin{equation}
    x_n = \sqrt{\bar{\alpha}_n}x_0 + \sqrt{1-\bar{\alpha}_n}\epsilon_a,
\end{equation}
where $\epsilon_a \in \mathcal{N}(\mu_a,\sigma_a^2\mathbf{I})$.

\subsection{Denoising Neural Network}
\begin{figure}
    \centering
    \includegraphics[width=0.99\linewidth]{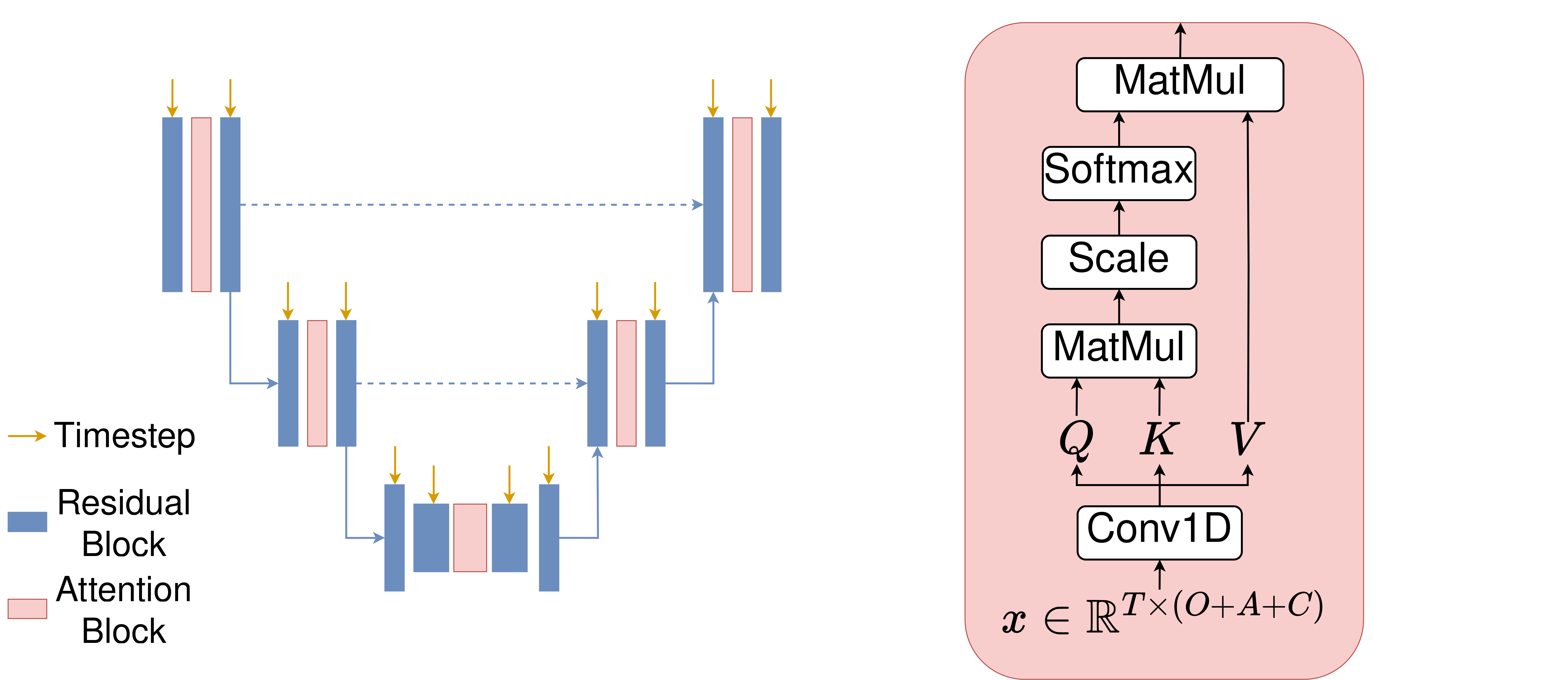}
    \caption{Architecture of the noise prediction neural network $\epsilon_\theta$. The network $\epsilon_\theta$ is based on U-Net and incorporates attention mechanisms.}
    \label{fig:Unet_attn}
\end{figure}

Following \cite{wang2023pdpp}, we use a U-Net as the noise prediction neural network $\epsilon_\theta$. 
To better learn the temporal dependencies between actions we further propose to incorporate an attention mechanism \cite{dhariwal2021diffusion, rombach2022high}.
\autoref{fig:Unet_attn} shows the architecture of the U-Net with attention. 
As indicated in \autoref{eq:unet}, the U-Net takes $x_n$ and noise schedule time step $n$ as input. 
$n$ is first passed to a time step block and then is fed to all residual blocks. 
The attention block takes $x$ as input and extracts Query ($Q$), Key ($K$) and Value ($V$) by a 1D convolutional layer and calculates the standard self-attention.

\subsection{Training}
In the training phase, the input $x_0$ to the diffusion model uses ground truth task class $c$.
However, during the inference phase, we need the predicted task class $\hat{c}$ since the procedure planning task has only access to $o_s$ and $o_g$ to infer $\pi$. Thus, we train an MLP for predicting $\hat{c}$ using $c$ as supervision. Next, we train the diffusion model using the following squared loss function,
\begin{equation}
    \mathcal{L} = \sum_{n=1}^N(\mu_\theta(x_n, n) - x_0)^2,
\end{equation}
where $\mu_\theta$ is the denoising neural network U-Net, $x_n$, $x_0$ and $n$ are the noised input, the input without noise and the timestep for adding noise respectively. 

\subsection{Inference}
\label{sec:inference}
\begin{figure}
    \centering
    \begin{subfigure}{0.14\textwidth}
         \centering
         \includegraphics[width=\textwidth]{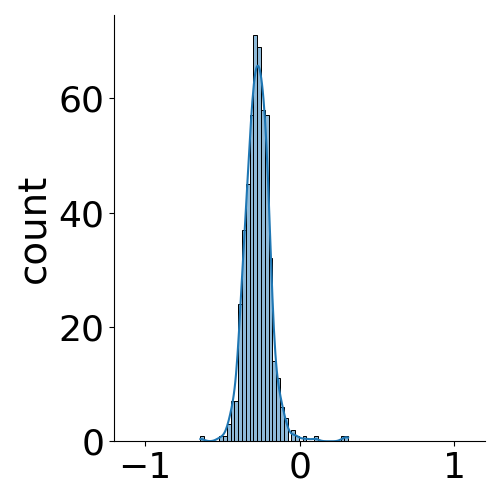}
         \caption{Crosstask}
    \end{subfigure}
    \begin{subfigure}{0.14\textwidth}
         \centering
         \includegraphics[width=\textwidth]{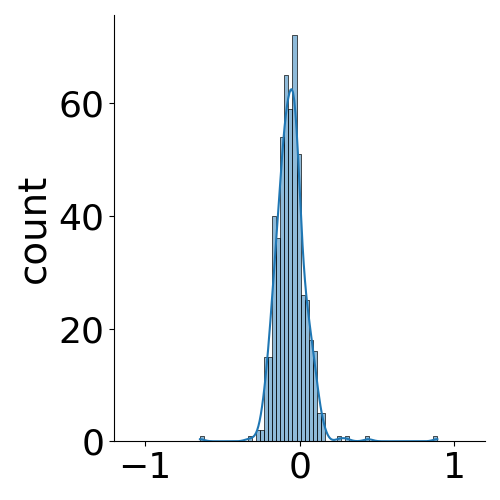}
         \caption{Coin}
    \end{subfigure}
    \begin{subfigure}{0.14\textwidth}
         \centering
         \includegraphics[width=\textwidth]{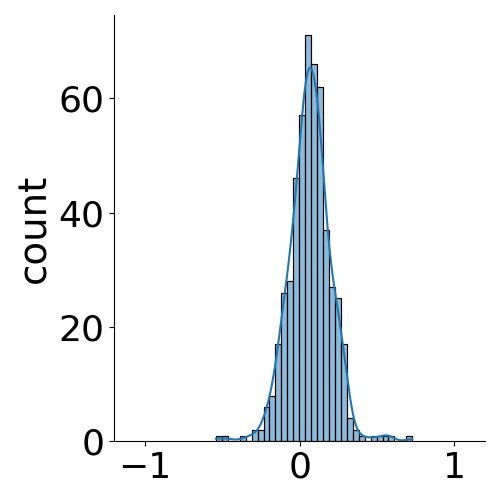}
         \caption{NIV}
    \end{subfigure}
    \caption{Examples of action embedding distributions from the CrossTask (a), Coin (b), and NIV datasets (c).}
    \label{fig:act_noise}
\end{figure}

During inference, only $o_s$, $o_g$ and the predicted task class $\hat{c}$ are given and we need to sample the action plan $\pi$. 
We start with constructing $x_n$. 
Then, the noise prediction neural network $\mu_\theta$ predicts the noise iteratively to denoise $x_n$ into $x_0$. Here, $x_n$ is
\begin{equation}\label{eq:inference_xn}
x_n = 
    \begin{bmatrix}
        \hat{c} & \hat{c}&  ... & \hat{c} & \hat{c} \\
        \epsilon_{a_1} & \epsilon_{a_2} & ... & \epsilon_{a_{T-1}} & \epsilon_{a_T} \\
        o_s & 0 & ... &0 & o_g
    \end{bmatrix},
\end{equation}
only the action plan $\pi$ is initialised with noise $\epsilon_{a_i}$ and $\epsilon_{a_i} \in \mathcal{N}(0, \sigma_{a_i}^2\mathbf{I})$ since the noise is only added to the actions during the noise-adding stage.
After $N$ steps of denoising, we take the action sequence $\hat{a}_{1:T}$ as the action plan $\pi$. In other words, all actions in $\pi$ are generated at once.
It is worth noting that we use action-aware noise masks in the noise-adding stage, the action embeddings are added to Gaussian noise (\autoref{eq:q_sample_act}). Since the action embeddings are approximately normally distributed (see \autoref{fig:act_noise}), the noised input after $N$ time steps is also approximately normally distributed with a different standard deviation compared to Gaussian noise (see \autoref{fig:noise}). Hence, instead of sampling from $\epsilon \in\mathcal{N}(0, \mathbf{I})$, we sample from $\epsilon_{a_i} \in \mathcal{N}(0, \sigma_{a_i}^2\mathbf{I})$. 
The comparison of using $\epsilon$ and $\epsilon_{a}$ for inference is shown in supplementary material.

%% file: Sections/experiments.tex
\section{Experiments}
\label{sec:exp}

\subsection{Datasets}

We evaluate \methodabbrev short on three instructional video benchmark datasets: CrossTask \cite{zhukov2019cross}, Coin \cite{tang2019coin}, and NIV \cite{alayrac2016unsupervised}.
The CrossTask dataset contains 2,750 videos, 18 tasks, and 105 action classes. 
The average number of actions per video is 7.6.
The Coin dataset has a much larger number of videos, tasks, and action classes:
In total, it contains 11,827 videos, 180 tasks, and 778 action classes.
The average number of actions per video is 3.6.
Finally, the NIV dataset is the smallest among the three datasets.
It contains 150 videos, five tasks, and 18 action classes as well as 9.5 actions per video on average.
We follow the data curation process used in previous work \cite{chang2020procedure, wang2023pdpp}:
For a video containing $m$ number of actions, we extract action sequences with the time horizon $T$ using a sliding window.
For an extracted action sequence $[a_i, ... , a_{i+T-1}]$, each action has a corresponding video clip.
We extract the video clip feature at the beginning of $a_i$ as the visual observation of the start state $o_s$ and the video clip feature at the end of $a_{i+T-1}$ as the goal state $o_g$.
To facilitate comparability, we use the pre-extracted features from previous work \cite{zhao2022p3iv, wang2023pdpp}.
The features were extracted by using a model \cite{miech2020end} which was pre-trained on the HowTo100M dataset \cite{miech2019howto100m}.
In addition to the video clip features, the model also generated embeddings for actions corresponding to video clips. 
We use the action embeddings for our action-aware diffusion.
We randomly use 70\% of the data for training and 30\% for testing.   

\subsection{Implementation Details}
We use AdamW \cite{loshchilov2017decoupled} as the optimizer for all datasets. For the training on the Crosstask dataset, we train with batch size 256 and 120 epochs. In each epoch, the train step is 200. The first 20 epochs are the warm-up stage, and the learning rate increases to $5e^{-4}$ linearly. In the last 30 epochs, the learning rate decays by 0.5 for every 5 epochs. For the Coin dataset, we train 800 epochs with batch size 256 with 200 train steps in each epoch. The warm-up state is 20 epochs where the learning rate increases to $1e^{-4}$ linearly. The learning rate decreases in the last 50 epochs with a decay rate 0.5. For the NIV dataset, we train 130 epochs with batch size 256. The train step in each epoch is 50. The learning rate increases to $1e^{-5}$ for $T=3$ and $3e^{-6}$ for $T=4$ in the first 90 epochs.

We use the same task predictor and adapt the same training strategy from \cite{wang2023pdpp} for predicting the task classes. The task predictor is a 4-layer MLP and it takes the image features of $o_s$ and $o_g$ as input and outputs the predicted task class $\hat{c}$. The accuracy is over 92\% on Crosstask, over 78\% on Coin and 100\% on NIV.

During inference, we sample $x_n$ using $\epsilon_a$ instead of $\epsilon$. The noised input approximately follows a normal distribution with a different mean and standard deviation other than the Gaussian noise.
The mean shifts and the standard deviation are smaller than one (see \autoref{fig:noise}).
We calculate the mean and standard deviations of the noised inputs on the training sets of Crosstask, Coin and NIV and use them for sampling in the inference on the test sets. The standard deviation of each action within the time horizon $T$ for all datasets is shown in \autoref{tab:std}.

\input{Figures/tab_std}

\subsection{Metrics}
We use three metrics for evaluation:
The first metric is the \textbf{Success Rate (SR)}.
An action plan is considered correct, and thus the procedure planning is a success, only if all actions in the plan are correct and the order of the actions is correct.
This is the most strict metric.
The second metric is the \textbf{mean Accuracy (mAcc)}.
The accuracy is calculated based on the individual actions in the action plan. The order of the actions is not considered. 
The last metric is \textbf{mean Single Intersection over Union (mSIoU)}. The calculation of IoU treats the predicted action plans and the ground truth action plans as sets and also does not consider the order of actions.
The works in \cite{chang2020procedure, zhao2022p3iv, bi2021procedure} calculated \textbf{mIoU} with all action plans in a mini-batch. 
However, as pointed out in \cite{wang2023pdpp}, this calculation is dependent on the batch size.
To ensure a fairer comparison, they proposed the \textbf{mSIoU} metric instead, which treats each single action plan as a set and is thus agnostic to the batch size. 
We also opted for the \textbf{mSIoU} metric. 

\subsection{Baselines}
We compare our method with several state-of-the-art baseline methods for procedure planning:
\begin{itemize}
    \item \textbf{DDN} \cite{chang2020procedure} uses MLPs and RNNs to learn the dynamics between actions and observations and uses search algorithms to sample the action plan.
    \item \textbf{Ext-GAIL} \cite{bi2021procedure} models action sequence as a Markov Decision Process (MDP) and uses imitation learning to learn policies to sample actions.
    \item \textbf{P$^3$IV} \cite{zhao2022p3iv} uses transformers with a memory block and a generative module trained with adversarial loss was used to sample the action plan.
    \item \textbf{PDPP} \cite{wang2023pdpp} uses a diffusion model to generate the action plan with the task class, start observation, and goal observation as conditions.
\end{itemize}

%% file: Figures/tab_std.tex
\begin{table}[H]\centering
\caption{Mean and standard deviations of $\epsilon_{a[1:T]}$ on the training sets of CrossTask, Coin and NIV.}\label{tab:std}
\resizebox{\linewidth}{!}{
\begin{tabular}{lcccccccccccc}\toprule
&  $T_1$ & & $T_2$& & $T_3$& & $T_4$& & $T_5$& & $T_6$\\
\cmidrule{2-13}
&$\mu$&$\sigma_{a}^2$&$\mu$&$\sigma_{a}^2$&$\mu$&$\sigma_{a}^2$&$\mu$&$\sigma_{a}^2$&$\mu$&$\sigma_{a}^2$&$\mu$&$\sigma_{a}^2$\\\cmidrule{2-13}
CrossTask &-0.27 &0.09 &-0.54 &0.13 & -0.81 &0.16 &-1.09 &0.18 & -1.35 &0.21 &-1.62 &0.22 \\
Coin &-0.04&0.59 &-0.08 &0.68 &-0.11 &0.72 &-0.14 &0.72 &- &-  &- &- \\
NIV & 0.06 &0.11 &0.12 &0.17 &0.19 &0.20 & 0.26 &0.23 &- &-  &- &- \\
%Ours &\textbf{} &\textbf{} & &\textbf{} \\
\bottomrule
\end{tabular}
}
\end{table}

%% file: Sections/results.tex
\section{Results}
\label{sec:results}
\input{Figures/tab_crosstask_T34}

\subsection{Action Embedding in Noise-Adding}
\begin{figure}[h]
    \centering 
    \includegraphics[width=0.45\textwidth]{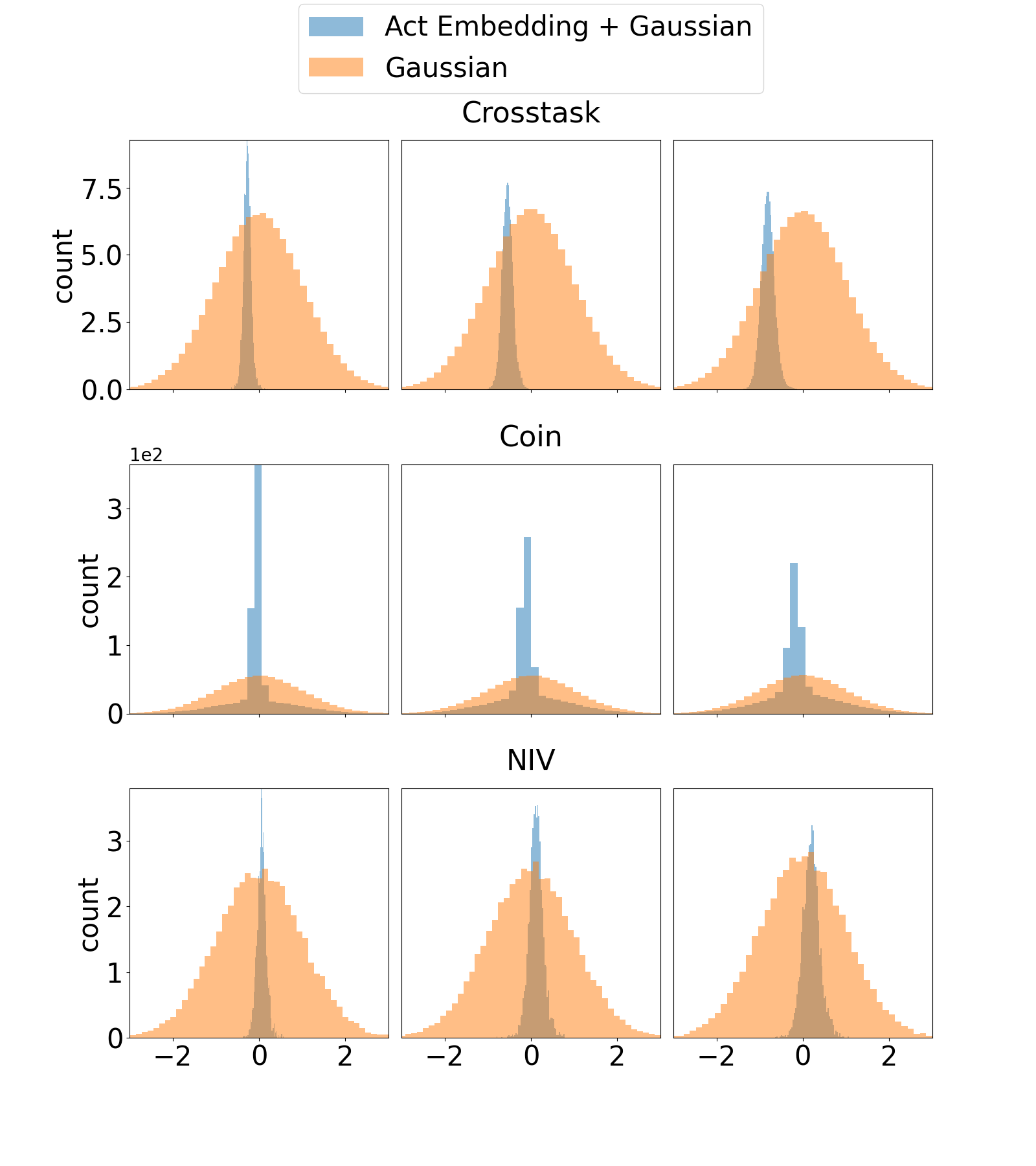}    
    \caption{The distributions of diffusion model input after $N$ steps of noise-adding for time horizon $T=3$. Each column shows the distributions at $a_i, i\in T$. The distribution in blue uses action embedding with Gaussian noise for noise-adding stage. The distribution in orange uses Gaussian noise only. The first row shows the distributions from CrossTask dataset. The second row shows the distributions from Coin dataset. The third row shows the distributions from NIV dataset.}
    \label{fig:noise}
\end{figure}
To verify if the noised input follows the normal distribution as mentioned in Section \ref{sec:inference}, we plot the noised actions $\epsilon_{a[1:T]}$ in $x_n$, i.e. the second row in \autoref{eq:inference_xn}, for time horizon $T=3$ on all datasets. Additionally, we plot the noised actions $\epsilon_{[1:T]}$ which use $\mathcal{N}(0, \mathbf{I})$ as the noise mask. The time step $N$ is 200, 200 and 50 for CrossTask, Coin and NIV respectively.
The results are shown in \autoref{fig:noise}. When using $\epsilon$ as the noise mask for the noise-adding, all noised actions approximately follow the same normal distribution for all three datasets. 
When using $\epsilon_a$ as the noise mask, we can observe from the figure that all action distributions in all datasets also follow normal distributions approximately. 
The mean of the distribution shifts towards the negative direction for CrossTask and Coin, and shifts toward the positive direction for NIV. For actions $a_2$ and $a_3$, the means shift further away from zero. The reason is that the mean of action embedding distribution is not zero (\autoref{fig:act_noise}), $a_3$ accumulates the action embeddings of $a_1$ and $a_2$. 
Additionally, the standard deviations of $\epsilon_a$ are increasing, i.e. $\sigma_{a3}^2>\sigma_{a2}^2, \sigma_{a2}^2>\sigma_{a1}^2$. The reason is that at $a_3$ all previous action embeddings are added to the noise mask.

\subsection{Comparison to SOTA Methods}
\label{subsec:result_sota}
\subsubsection{CrossTask}
\autoref{tab:result_crosstask_T34} shows the results on the CrossTask dataset with time horizon $T=3$ and $T=4$. \methodabbrev short achieves SOTA performances for all metrics in both time horizons. Note that DNN, Ext-GAIL, and P$^3$IV only reported mIoU, and only PDPP reported mSIoU.  
We also evaluate our method with the longer time horizon $T\in \{3,4,5,6\}$. \autoref{tab:result_crosstask_T3456} shows the SR for all the time horizons. We again have SOTA performances in all time horizons. 
\input{Figures/tab_crosstask_T3456}

\subsubsection{Coin}
\input{Figures/tab_niv_coin}

\autoref{tab:results_niv_coin} shows the results on the Coin dataset with time horizon $T=3$ and $T=4$. For $T=3$, we achieved SOTA performance on SR and mSIOU. Although the mAcc is slightly lower than PDPP, our SR is still 2.67\% higher than PDPP. This means that \methodabbrev short can learn the temporal dependencies better than PDPP even though the percentage of correctly predicted actions (ignoring the order) is slightly lower. 
When $T=4$, we have SOTA results on all metrics. Our mAcc is slightly better than PDPP, SR and mSIoU are 4.63\% and 4.84\% higher. This also shows our method is better at capturing the temporal dependencies.

\subsubsection{NIV}
The results on the NIV dataset are shown in \autoref{tab:results_niv_coin}. We obtain SOTA performance on all metrics in both time horizons. The SR is 2.76\% higher than PDPP when $T=3$ and 2.59\% higher when $T=4$. The mACC is 1.18\% and 1.39\% higher when $T=3$ and $T=4$. The mSIoU is also slightly higher when $T=3$ and $T=4$.

\subsection{Ablation Study}

\subsubsection{Noise Mask}
\label{subsec:result_noise_mask}
In \autoref{eq:multi-add_mask}, we describe the $\textbf{MultiAdd}$ action-aware noise mask. At each time step within the time horizon $T$, we accumulate the embeddings of all previous actions. We want to compare it with $\textbf{SingleAdd}$ noise mask described as follows, 
\begin{equation}
    \begin{bmatrix}
        0 & 0 & ... & 0 \\
        g(a_{e_{1}}) & g(a_{e_{2}}) & ...  & g(a_{e_{T}}) \\
        0 & 0 & ... &0 
    \end{bmatrix},
\end{equation}
where at each time step only one action embedding is added to the noise mask. 

\begin{figure}
    \centering
    \includegraphics[width=0.99\linewidth]{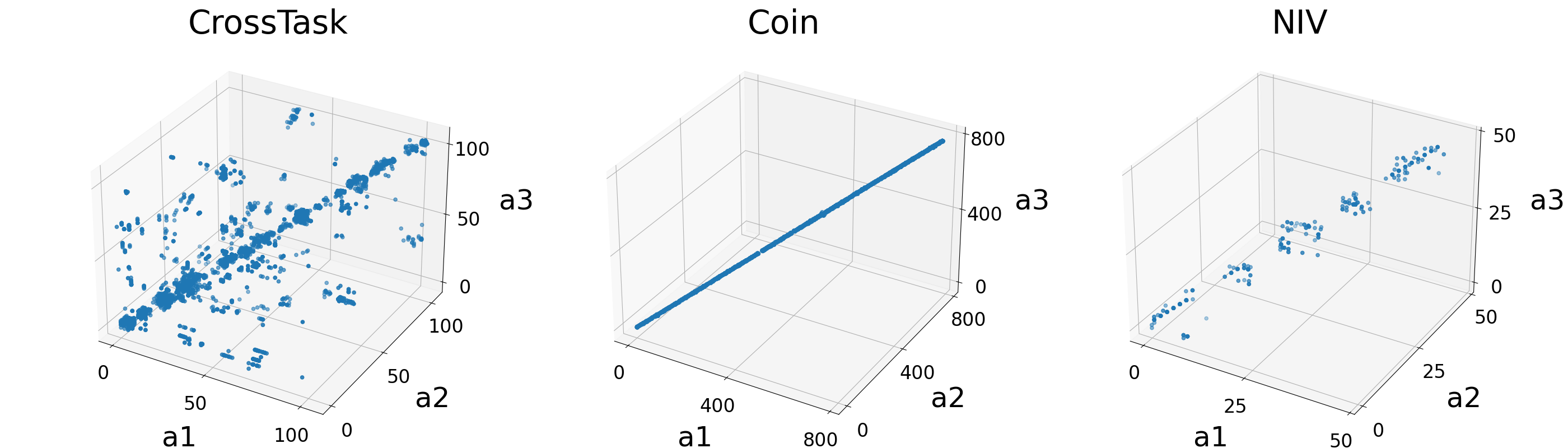}
    \caption{Distributions of action labels for $T=3$. Each point in the 3D coordinate represents one action sequence. The coordinates in x, y and z directions is the action labels for $a_1$, $a_2$ and $a_3$ respectively.}
    \label{fig:dist_gt_action}
\end{figure}

\autoref{tab:single_ablation} shows the results of the MultiAdd mask, SingleAdd mask and without mask (NoMask) on all three datasets with time horizon $T=3$ and $T=4$. 
MultiAdd outperforms SingleAdd and NoMask on all metrics on the Coin dataset and NIV dataset for both time horizons. 
On the CrossTask dataset, SingleAdd performs the best on SR when $T=3$. NoMask performs the best on mAcc and MSIoU for $T=3$ and all metrics for $T=4$. Although NoMask has the best overall performance on CrossTask, the results of MultiAdd, SingleAdd and NoMask are comparable. Additionally, the performances of NoMask are worse on Coin and NIV, especially on Coin. 
We interpret the reason for the performance differences is that the differences in action label compositions in action sequences from datasets. \autoref{fig:dist_gt_action} shows the distributions of ground truth action labels in action sequences.
The action labels in action sequences in CrossTask are more scattered than in Coin and NIV. For instance, lots of action sequences look like $a_{1:T} = [56,0,57]$, where 56, 0 and 57 are the action labels. And most of the action sequences in Coin and NIV look like $a_{1:T} = [48,49,50]$. Overall the distribution of action sequences is more linear in Coin and NIV. We think this is the reason that the action-aware mask works better since it is easier for the action-aware mask to build temporal dependencies. 
Overall, MultiAdd can perform better when the distribution of action labels is more linear (Coin and NIV). For the more scattered distributions, MultiAdd and SingleAdd are not as effective. Nevertheless, MultiAdd still achieves the SOTA performances. 

\input{Figures/tab_single_ablation}

\subsubsection{Self-Attention in U-Net}
\label{subsec:result_unet}
\input{Figures/tab_attn_ablation}

We study the effect of the self-attention mechanism in the U-Net in this section. We use the Multi-Add noise mask and test the U-Net with and without self-attention on all three datasets. The results are shown in \autoref{tab:ablation_attn}. 
\methodabbrev short with and without self-attention achieve comparable results on the CrossTask dataset. 
On the Coin dataset, \methodabbrev short with self-attention performs better on all metrics when $T=3$. \methodabbrev short without self-attention performs better on all metrics when $T=4$, although the results of \methodabbrev short with and without self-attention are comparable.
On the NIV dataset, \methodabbrev short with self-attention outperforms the one without self-attention on all metrics with both time horizons. 
Overall, \methodabbrev short with self-attention performs better than without self-attention on the NIV dataset, while they have comparable performance on the other two datasets. We interpret the reason for this as the size of NIV is much smaller than the other two. \methodabbrev short with attention can better learn the temporal dependencies integrated into the noise mask when the data is limited.

%% file: Figures/tab_crosstask_T34.tex
\begin{table*}
\centering
\caption{Results on CrossTask dataset with time horizon $T=3$ and $T=4$. Numbers in \textbf{Bold} indicate the best results. The arrow ↑ means higher numbers are better.}\label{tab:result_crosstask_T34}
%\scriptsize
\resizebox{0.8\linewidth}{!}{
\begin{tabular}{lcccccc}\toprule
&T=3 &  & &T=4 &  & \\\cmidrule{2-7}
 &SR(\%)↑ &mAcc(\%)↑ &mSIoU(\%)↑ &SR(\%)↑ &mAcc(\%)↑  &mSIoU(\%)↑ \\\midrule
DDN \cite{chang2020procedure} &12.18 &31.29 &- &5.97 &27.10  &- \\
Ext-GAIL \cite{bi2021procedure} &21.27 &49.46 &- &16.41 &43.05 &- \\
P$^3$IV \cite{zhao2022p3iv} &23.34 &49.96 &- &13.40 &44.16 &- \\
PDPP \cite{wang2023pdpp} &37.20 &64.67 &66.57 &21.48 &57.82 &65.13 \\
\methodabbrev short-MultiAdd (Ours) &\textbf{37.79} &\textbf{65.38} &\textbf{67.45} &\textbf{22.43} &\textbf{59.42} &\textbf{66.04} \\
\bottomrule
\end{tabular}
}
\end{table*}

%% file: Figures/tab_crosstask_T3456.tex
\begin{table}[hbt!]\centering
\caption{Results on CrossTask dataset with time horizon $T \in \{3,4,5,6\}$. Numbers in \textbf{Bold} indicate the best results. The arrow ↑ means higher numbers are better.}\label{tab:result_crosstask_T3456}
\resizebox{\linewidth}{!}{
\begin{tabular}{lccccc}\toprule
&T=3 &T=4 &T=5 &T=6 \\\cmidrule{2-5}
&SR(\%)↑ &SR(\%)↑ &SR(\%)↑ &SR(\%)↑ \\\midrule
DDN \cite{chang2020procedure} &12.18 &5,97 &3.10 &1.20 \\
Ext-GAIL \cite{bi2021procedure} &21.27 &16.41 &- &- \\
P$^3$IV \cite{zhao2022p3iv} &23.34 &13.40 &7.21 &4.40 \\
PDPP \cite{wang2023pdpp} &37.20 &21.48 &13.45 &8.41 \\
\methodabbrev short-MultiAdd (Ours) &\textbf{37.79} &\textbf{22.43} &\textbf{13.89} &\textbf{9.66} \\
\bottomrule
\end{tabular}
}
\end{table}

%% file: Figures/tab_niv_coin.tex
\begin{table*}[hbt!]\centering
\caption{Results on Coin and NIV datasets with time horizon $T=3$ and $T=4$. Numbers in \textbf{Bold} indicate the best results. The arrow ↑ means higher numbers are better. \methodabbrev short means our method using MultiAdd mask.}\label{tab:results_niv_coin}
\scriptsize
\begin{tabular}{lccccccc}\toprule
& &COIN & & &NIV & & \\\cmidrule{3-8}
Horizon &Models &SR(\%)↑ &mAcc(\%)↑ &mSIoU(\%)↑ &SR(\%)↑ &mAcc(\%)↑ &mSIoU(\%)↑ \\\midrule
T=3 &DDN \cite{chang2020procedure} &13.90 &20.19 &- &18.41 &32.54 &- \\
&Ext-GAIL \cite{bi2021procedure} &- &- &- &22.11 &42.20 &- \\
&P$^3$IV \cite{zhao2022p3iv} &15.40 &21.67 &- &24.68 &49.01 &- \\
&PDPP \cite{wang2023pdpp} &21.33 &\textbf{45.62} &51.82 &30.20 &48.45 &57.28 \\
&\methodabbrev short-MultiAdd (Ours) &\textbf{24.00} &45.42 &\textbf{54.29} &\textbf{32.96} &\textbf{49.26} &\textbf{57.84} \\
\midrule
T=4 &DDN \cite{chang2020procedure} &11.13 &17.71 &- &15.97 &2.73 &- \\
&Ext-GAIL \cite{bi2021procedure} &- &- &- &19.91 &36.31 &- \\
&P$^3$IV \cite{zhao2022p3iv} &11.32 &18.85 &- &20.14 &28.36 &- \\
&PDPP \cite{wang2023pdpp} &14.41 &44.10 &51.39 &26.67 &46.89 &59.45 \\
&\methodabbrev short-MultiAdd (Ours) &\textbf{18.04} &\textbf{44.54} &\textbf{56.23} &\textbf{29.26} &\textbf{48.14} &\textbf{60.71} \\
\bottomrule
\end{tabular}
\end{table*}

%% file: Figures/tab_single_ablation.tex
\begin{table*}[hbt!]\centering
\caption{Comparison between MultiAdd noise mask, SingleAdd noise mask and without mask on Crosstask, Coin and NIV datasets. Numbers in \textbf{Bold} indicate the best results. The arrow ↑ means higher numbers are better.}\label{tab:single_ablation}
\scriptsize
%\resizebox{\linewidth}{!}{
\begin{tabular}{lccccccc}\toprule
& &T=3 & & &T=4 & & \\\cmidrule{3-8}
Dataset &Models &SR↑ &mAcc↑ &mSIoU↑ &SR↑ &mAcc↑ &mSIoU↑ \\\midrule
\multirow{2}{*}{CrossTask} &\methodabbrev short-Multi &37.79 &65.38 &67.45 &22.43 &59.42 &66.04 \\ 
&\methodabbrev short-Single &\textbf{38.21} &65.34 &67.25 &22.32 &58.92 &65.26 \\
&\methodabbrev short-NoMask &37.92 &\textbf{65.53} &\textbf{67.65} &\textbf{22.99} &\textbf{59.48} &\textbf{66.23} \\\midrule
\multirow{2}{*}{Coin} &\methodabbrev short-Multi &\textbf{24.00} &\textbf{45.42} &\textbf{54.29} &\textbf{18.04} &\textbf{44.54} &\textbf{56.23} \\
&\methodabbrev short-Single &21.52 &43.13 &52.98 &14.97 &42.56 &55.04 \\
&\methodabbrev short-NoMask &12.46 &36.08 &43.44 &2.47 &26.96 & 34.70 \\\midrule
\multirow{2}{*}{NIV} &\methodabbrev short-Multi &\textbf{32.96} &\textbf{49.26} &\textbf{57.84} &\textbf{29.26} &\textbf{48.14} &\textbf{60.71} \\
&\methodabbrev short-Single &30.74 &47.03 &56.00 &25.76 &44.98 &57.92 \\
&\methodabbrev short-NoMask &29.26 &44.81 &54.49 &21.40 &35.59 &51.38 \\\midrule
\bottomrule
\end{tabular}
%}
\end{table*}

%% file: Figures/tab_attn_ablation.tex
\begin{table*}[hbt!]\centering
\caption{Comparison between the U-Net with self-attention (w attention) and the U-Net without self-attention (w/o attention). Numbers in \textbf{Bold} indicate the best results. The arrow ↑ means higher numbers are better.}\label{tab:ablation_attn}
\scriptsize
%\resizebox{\linewidth}{!}{
\begin{tabular}{lccccccc}\toprule
& &T=3 & & &T=4 & & \\\cmidrule{3-8}
Dataset &Models &SR↑ &mAcc↑ &mSIoU↑ &SR↑ &mAcc↑ &mSIoU↑ \\\midrule
\multirow{2}{*}{CrossTask} &w attention &\textbf{37.79} &65.38 &\textbf{67.45} &22.43 &\textbf{59.42} &66.04 \\
&w/o attention &37.75 &\textbf{65.47} &\textbf{67.45} &\textbf{22.56} &59.17 &\textbf{66.15} \\ \midrule
\multirow{2}{*}{Coin} &w attention &\textbf{24.00} &\textbf{45.42} &\textbf{54.29} &18.04 &44.54 &56.23 \\
&w/o attention &22.88 &44.52 &53.56 &\textbf{18.36} &\textbf{44.88} &\textbf{56.49} \\ \midrule
\multirow{2}{*}{NIV} &w attention &\textbf{32.96} &\textbf{49.26} &\textbf{57.84} &\textbf{29.26} &\textbf{48.14} &\textbf{60.71} \\
&w/o attention &32.22 &48.52 &57.58 &28.82 &46.07 &59.12 \\
\bottomrule
\end{tabular}
%}
\end{table*}

%% file: Sections/conclusion.tex
\section{Conclusion}
\label{sec:conclusion}

In this work, we propose \methodabbrev short, an action-aware diffusion model, to tackle the challenge of procedure planning in instructional videos. We integrate temporal dependencies between actions by adding action embeddings to the noise mask during the diffusion process.
We achieve SOTA performances on three procedure planning datasets across multiple metrics, showing the novelty of adding action embedding in the noise mask as the modelling of temporal dependencies.